\documentclass[10pt,journal,compsoc]{IEEEtran}

\ifCLASSOPTIONcompsoc
  \usepackage[nocompress]{cite}
\else
  \usepackage{cite}
\fi

\usepackage{caption}
\usepackage{graphicx, subfig}
\usepackage{color}
\usepackage{amsmath,amssymb,amsthm}
\usepackage{bm}
\usepackage{stfloats}
\usepackage{epsfig}
\usepackage{multirow}

\usepackage{makecell}
\usepackage{url}

\renewcommand{\mathbf}{\boldsymbol}

\begin{document}
\begin{sloppypar}

\title{CAD-PU: A Curvature-Adaptive Deep Learning Solution for Point Set Upsampling}

\author{Jiehong Lin, Xian Shi, Yuan Gao, Ke Chen, Kui Jia
\IEEEcompsocitemizethanks{\IEEEcompsocthanksitem J. Lin, X. Shi, K. Chen and K. Jia are with the School of Electronic and Information Engineering, South China University of Technology, Guangzhou, China.\protect\\
Emails: lin.jiehong@mail.scut.edu.cn, ausxian@mail.scut.edu.cn, chenk@scut.edu.cn, kuijia@scut.edu.cn

\IEEEcompsocthanksitem Y. Gao is with Tencent AI Lab, Shenzhen, China.\protect\\
Email: ethan.y.gao@gmail.com
\IEEEcompsocthanksitem K. Jia is the corresponding author.
}}

\IEEEtitleabstractindextext{%

\begin{abstract}
Point set is arguably the most direct approximation of an object or scene surface, yet its practical acquisition often suffers from the shortcoming of being noisy, sparse, and possibly incomplete, which restricts its use for a high-quality surface recovery. Point set upsampling aims to increase its density and regularity such that a better surface recovery could be achieved. The problem is severely ill-posed and challenging, considering that the upsampling target itself is only an approximation of the underlying surface. Motivated to improve the surface approximation via point set upsampling, we identify the factors that are critical to the objective, by pairing the surface approximation error bounds of the input and output point sets. It suggests that given a fixed budget of points in the upsampling result, more points should be distributed onto the surface regions where local curvatures are relatively high. To implement the motivation, we propose a novel design of Curvature-ADaptive Point set Upsampling network (CAD-PU), the core of which is a module of curvature-adaptive feature expansion. To train CAD-PU, we follow the same motivation and propose geometrically intuitive surrogates that approximate discrete notions of surface curvature for the upsampled point set. We further integrate the proposed surrogates into an adversarial learning based curvature minimization objective, which gives a practically effective learning of CAD-PU. We conduct thorough experiments that show the efficacy of our contributions and the advantages of our method over existing ones. Our implementation codes are publicly available at \url{https://github.com/JiehongLin/CAD-PU}.
\end{abstract}

}

\maketitle

\IEEEpeerreviewmaketitle

\IEEEraisesectionheading{
\section{Introduction}\label{sec:introduction}}

\IEEEPARstart{S}{urface} modeling of 3D objects or scenes is one of the fundamental problems in vision and graphics. Due to its nature of continuity and high-order smoothness, a surface is usually represented as various approximations, with point set, volume, and mesh as the prominent examples. With the popularity of various 3D sensors and the development of surface reconstruction via techniques of multi-view stereo \cite{kar2017learning, yao2018mvsnet}, point sets become the most direct representations acquired from real-world sensing. However, such practically captured point sets tend to be noisy, sparse, and possibly incomplete, which restricts their use for high-quality recovery of the underlying surfaces. \emph{Point set upsampling} aims to improve the quality by producing a denser and more regular point set result \cite{PUNet,MPU,PUGAN}, such that it serves as a better approximation of the underlying surface.

Point set upsampling is by nature an ill-posed problem. Given an input, sparse point set as a poor surface approximation, there exist infinitely numerous solutions whose subsamplings produce the same input. Compared with its 2D counterparts, e.g., image super-resolution \cite{SRBookCollectionByMilanfar}, the problem of point set upsampling is even more challenging considering that the target is again a surface approximation whose realization is less well defined, which makes it less obvious to specify the upsampling objectives. As such, existing methods take different criteria to achieve the goal. For example, a recent line of deep learning research \cite{PUNet,PUGAN} promotes a uniform distribution of points on the surface, and designs corresponding network modules to meet the criterion; more classical results derive shape operators on the discrete point set via modeling of surface approximation, where curvature-related criteria are often suggested \cite{MLS,Lange05anisotropicsmoothing}. While local regularity, e.g., uniformity, is a nice property of discrete point sets, it is how the points globally distribute on the surface that determines the quality of geometric and topological surface approximations \cite{Levin98theapproximation,Gu07GeomModeling}.

In this work, we are motivated to develop deep learning solutions of point set upsampling by formulating it as a problem of learning to improve surface approximation. Following the analyzing framework of moving least squares (MLS) \cite{MLS}, we identify factors critical to the objective by pairing the surface approximation error bounds of the input and output point sets. Given a fixed budget of points in the upsampling result, the analysis suggests that more points should be distributed onto high-curvature surface areas in order for the result to be optimal for surface approximation. To implement the motivation, we propose in this work novel designs hinging on approximate versions of \emph{mean curvature} defined at local surface patches. Specifically, we propose Curvature-ADaptive Point set Upsampling network (CAD-PU) whose key design is a module of \emph{curvature-adaptive feature expansion}. The proposed module is able to expand point-wise features in a globally curvature-adaptive manner, while simultaneously maintaining the local regularity implicitly; as a result, high-quality upsamplings are made possible by the subsequent layers of geometric feature learning and regression of upsampled points. To train CAD-PU, we follow the same motivation of learning to improve surface approximation, and propose geometrically intuitive surrogates that approximate discrete notions of surface curvature for the upsampled point set. We further integrate the proposed surrogates into an adversarial learning based curvature minimization objective, which yields a practically effective training of CAD-PU. We conduct ablation studies that confirm the efficacy of our novel designs. Comparative experiments with existing methods, including those of optimization \cite{huang2013edge} and learning based approaches \cite{PUNet,MPU,PUGAN}, show the quantitative and qualitative advantages of our method in terms of supporting high-quality surface recovery via point set upsampling. Experiments with noisy inputs and real scans also demonstrate the robustness of our method.

\section{Related Works}
\label{SecRelatedWork}

\noindent \textbf{Optimization of point set upsampling} Optimization based methods are paired with the development of point set surface modeling and approximation. They typically define shape operators by which surface recovery and/or point set upsampling can be achieved. For example, built on the MLS projection, Alexa \textit{et al.} \cite{MLS} propose to upsample a point set by computing the Voronoi diagram on the MLS surface and inserting new points at the vertices of this diagram. Lipman \textit{et al.} \cite{lipman2007parameterization} introduce a different operator of locally optimal projection (LOP) for point resampling and surface approximation. Huang \textit{et al.} \cite{huang2009consolidation} further develop a weighted LOP in order to generate noise- and outlier-free point sets, and an edge-aware point set resampling (EAR) method \cite{huang2013edge} that first resamples point sets away from edges and then progressively approaches the edge singularities. For point set denoising and completion, Wu \textit{et al.} \cite{wu2015deep} propose a new representation by associating each surface point with an inner point residing on the meso-skeleton, and consequently, optimization benefits from global geometry constraints from the meso-skeleton.

\vspace{0.1cm}
\noindent \textbf{Deep learning point set upsampling} More recently, deep representation learning of point sets is proposed for both semantic \cite{qi2017pointnet, wang2019dynamic} and generative tasks \cite{achlioptas2018learning}, including those for point set upsampling \cite{PUNet,MPU,PUGAN} and completion \cite{yuan2018pcn}. Yu \textit{et al.} \cite{PUNet} firstly propose a data-driven solution of PU-Net that exploits multi-branch convolutions to expand multi-level point features for prediction of point set upsampling. Wang \textit{et al.} \cite{MPU} present MPU, a patch-based progressive upsampling network that learns different levels of details in multiple steps, resulting in improved upsampling. Li \textit{et al.} \cite{PUGAN} make use of the power of generative adversarial networks (GAN) and propose PU-GAN, which sets the existing state of the art for deep learning based point set upsampling.

The present work follows the recent success of deep learning solutions, and improve both the architectural designs and learning objectives such that they are more aligned with the classical results of point set surface approximation \cite{Levin98theapproximation}. Experiments show that our connection of classical and modern strategies sets the new state of the art.

\section{Point Set Upsampling for an Improved Approximation of the Underlying Surface}
\label{SecMotivation}

Assume a training set $\{\mathcal{P}_i, \mathcal{Q}_i^{*}\}_{i=1}^M$ of $M$ sample pairs, where each $\mathcal{P}=\{\bm{p}_i  \in \mathbb{R}^3\}_{i=1}^N$ represents an input, possibly sparse, point set and $\mathcal{Q}^{*}=\{\bm{q}_i^{*} \in \mathbb{R}^3\}_{i=1}^{rN}$ represents the ground-truth output whose number of points is $r$ factors of that of $\mathcal{P}$. In this work, we consider a learning task of point set upsampling that aims to learn a generator $G(\cdot)$, e.g., a deep network, such that given any test $\mathcal{P}$ sampled from an underlying surface $\mathcal{S}^{*}$, the generator produces $\mathcal{Q} = G(\mathcal{P})$ that can better approximate $\mathcal{S}^{*}$.

For any pair $\{\mathcal{P}, \mathcal{Q}\}$, assume that their underlying surface $\mathcal{S}^{*}$ is $C^{\infty}$ smooth. By differential geometry we know that $\mathcal{S}^{*}$ can be locally represented as functions defined on coordinate systems corresponding to its local surface patches. Following the analyzing framework of moving least squares (MLS) \cite{MLS}, when using polynomials as the local functions over 2D domains centered at individual points $\{ \bm{q} \in \mathcal{Q} \}$ and when the local domains are properly defined \cite{MLS}, we have $\mathcal{S}_{\mathcal{Q}} \in C^{\infty}$ as an approximation of $\mathcal{S}^{*}$, where $\mathcal{S}_{\mathcal{Q}}$ denotes a surface reconstructed from the point set $\mathcal{Q}$; we correspondingly write as $\mathcal{S}_{\bm{q}}$ for the local, polynomial approximation of the surface patch centered at any $\bm{q} \in \mathcal{Q}$. We similarly define $\mathcal{S}_{\mathcal{P}} \in C^{\infty}$ as another approximation of $\mathcal{S}^{*}$, and $\{ \mathcal{S}_{\bm{p}} \}$ contains its local approximations. The objective of point set upsampling is thus to produce $\mathcal{Q} = G(\mathcal{P})$ such that $\mathcal{S}_{\mathcal{Q}}$ constructed from an increased number of points improves the approximation of $\mathcal{S}^{*}$ over $\mathcal{S}_{\mathcal{P}}$, i.e., $E(\mathcal{S}_{\mathcal{Q}}, \mathcal{S}^{*}) \leq E(\mathcal{S}_{\mathcal{P}}, \mathcal{S}^{*})$.

Assume that the polynomial functions are of degree $n$. From \cite{Levin98theapproximation} we know that approximating a local surface patch with $\mathcal{S}_{\bm{q}}$ gives the error bound $\| \mathcal{S}_{\bm{q}} - \mathcal{S}_{\bm{q}}^{*} \| \leq C(\| \mathcal{S}^{*(n+1)}_{\bm{q}} \|) \cdot \omega_{\bm{q}}^{n+1}$ , where $\omega_{\bm{q}}$ denotes the width of local 2D domain centered at $\bm{q}$ and $C$ is a constant depending on the $(n+1)^{th}$ local surface derivatives. When $\{\mathcal{S}_{\bm{q}_i} \}_{i=1}^{rN}$ cover the surface with no holes, we have that $\mathcal{S}_{\mathcal{Q}}$, as the union of $\{\mathcal{S}_{\bm{q}_i} \}_{i=1}^{rN}$, approximates $\mathcal{S}^{*}$ with the following error bound
\vspace{-0.05cm}
\begin{equation}\label{EqnErrBoundQ}
E(\mathcal{S}_{\mathcal{Q}}, \mathcal{S}^{*}) = \sum_{i=1}^{rN} \| \mathcal{S}_{\bm{q}_i} - \mathcal{S}^{*}_{\bm{q}_i} \| \leq \sum_{i=1}^{rN} C(\| \mathcal{S}^{*(n+1)}_{\bm{q}_i} \|) \cdot \omega_{\bm{q}_i}^{n+1} .
\end{equation}
\vspace{-0.05cm}
We similarly have the following error bound for $\mathcal{S}_{\mathcal{P}}$
\vspace{-0.05cm}
\begin{equation}\label{EqnErrBoundP}
E(\mathcal{S}_{\mathcal{P}}, \mathcal{S}^{*}) = \sum_{i=1}^N \| \mathcal{S}_{\bm{p}_i} - \mathcal{S}^{*}_{\bm{p}_i} \| \leq \sum_{i=1}^N C(\| \mathcal{S}^{*(n+1)}_{\bm{p}_i} \|) \cdot \omega_{\bm{p}_i}^{n+1} .
\end{equation}
\vspace{-0.05cm}
The bounds (\ref{EqnErrBoundQ}) and (\ref{EqnErrBoundP}) tell that approximation errors depend on a coupled factor of local surface derivatives and widths of the corresponding local 2D domains. When $n=1$, i.e., $\mathcal{S}_{\mathcal{Q}}$ and $\mathcal{S}_{\mathcal{P}}$ are piecewise linear approximations of $\mathcal{S}^{*}$, the bounds are directly relevant to local surface curvatures. We consider this case in the following analysis. Note that it is quite reasonable to set $n=1$ to have piecewise linear surface approximations, since our visual system is insensitive to surface smoothness beyond second order \cite{Marr82}.

By comparing (\ref{EqnErrBoundQ}) and (\ref{EqnErrBoundP}) we have that, under the condition that the residing of upsampled $\mathcal{Q} = \{ \bm{q}_i \}_{i=1}^{rN}$ on $\mathcal{S}^{*}$ follows the same distribution as that of $\mathcal{P} = \{ \bm{p}_i \}_{i=1}^N$, point set upsampling reduces the approximation error since, on average, $\omega_{\bm{q}} \leq \omega_{\bm{p}}$. To further reduce the approximation error via point set upsampling, we are motivated by the following two observations:
\vspace{-0.18cm}
\begin{enumerate}
\item given a fixed budget of $rN$ points in the upsampled $\mathcal{Q}$, distributing more points to surface patches at $\{\bm{q}\}$ that have higher values of curvature (and thus higher values of $\{ C(\| \mathcal{S}^{*(2)}_{\bm{q}} \|) \}$) reduces the corresponding widths $\{\omega_{\bm{q}}\}$, and thus reduces the overall RHS summation in the bound (\ref{EqnErrBoundQ});

\item given that the underlying surface $\mathcal{S}^{*}$ for a test $\mathcal{P}$ is unknown, an optimal $\mathcal{S}^{*}$ with smaller summation of local curvatures $\sum_{i=1}^{rN} \| \mathcal{S}^{*(2)}_{\bm{q}_i} \| $ achieves a lower error bound (\ref{EqnErrBoundQ}), which is also aligned with the desired property of \emph{fair} surface \cite{botsch2010polygon} \footnote{The fairness of a surface is related to its aesthetic perception. There is no a unique definition of surface fairness; a surface is generally considered \emph{fair} if its curvatures or variations of curvature are globally minimized \cite{botsch2010polygon}. }.
\end{enumerate}
\vspace{-0.1cm}
To implement the first observation, we propose \emph{Curvature-Adaptive Feature Expansion} as the key module in our proposed generator of \emph{Curvature-ADaptive Point set Upsampling network (CAD-PU)}, which is specifically designed to distribute more points to high-curvature surface patches during the point set upsampling process. For the second one, we propose an objective to train CAD-PU by minimizing discrete and approximate surrogates of surface curvatures for the upsampled point set. These technical designs constitute the main contributions of our proposed CAD-PU, whose details are presented shortly.

\vspace{-0.25cm}
\section{The Proposed Method}
\label{SecMethod}

\begin{figure*}[!t]
\centering
\vspace{-0.1cm}
\includegraphics[width=0.9\textwidth]{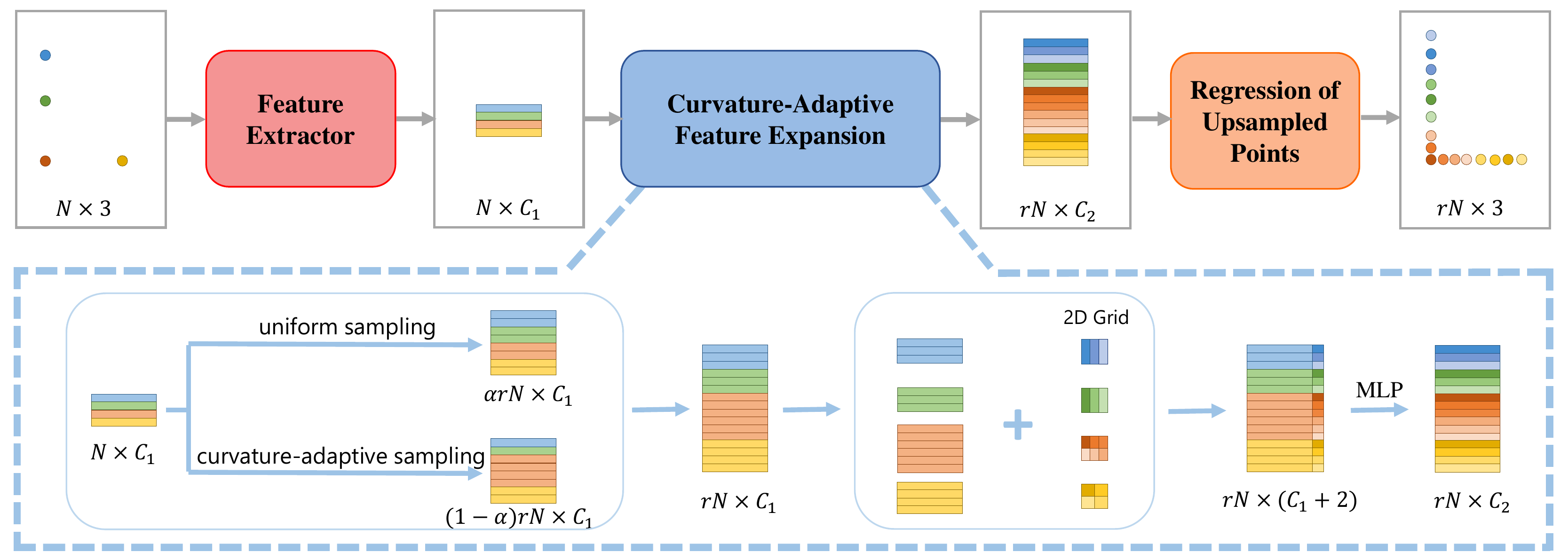}
\vspace{-0.1cm}
\caption{Architecture of our proposed Curvature-ADaptive Point set Upsampling network (CAD-PU). Given an input set of $N$ points, the network first extracts point-wise features using a same feature extraction module as in \cite{MPU}. It then expands the $N$ point-wise features as $rN$ ones using our proposed module of curvature-adaptive feature expansion. Coordinates of the upsampled points are finally obtained by a simple regression. The key of curvature-adaptive feature expansion is a hybrid scheme that combines uniform and curvature-adaptive samplings of point-wise features for expansion. Discrimination among the sampled, possibly duplicate, features is also injected by augmenting the features with 2D coordinates at corners of regular grids defined on an arbitrary but fixed 2D plane; this is necessary to make the upsampled points spatially spread out. We train CAD-PU using a novel objective aiming for an improved approximation of the underlying surface via point set upsampling. For better illustration, different points and their associated feature vectors are coded with different colors.}
\label{FigNetwork}
\vspace{-0.4cm}
\end{figure*}

In this section, we first present the elements of our proposed Curvature-ADaptive Point set Upsampling network (CAD-PU), whose illustration is given in Fig. \ref{FigNetwork}. We then introduce our learning objectives that train CAD-PU to generate upsampled point sets.

\vspace{-0.35cm}
\subsection{Curvature-Adaptive Point-Upsampling Network}
\label{NetworkArchitecture}

CAD-PU processes an input point set by independently and in parallel upsampling its point subsets corresponding to local surface patches. Without loss of generality, we again use $\mathcal{P} = \{\bm{p}_i \in \mathbb{R}^3\}_{i=1}^N$ to denote an input point subset, whose coordinates are fed into a \emph{feature extractor} to learn point-wise features $\mathcal{F}^1 = \{\bm{f}^1_i \in \mathbb{R}^{C_1} \}_{i=1}^{N}$. Our proposed module of \emph{curvature-adaptive feature expansion} then learns from $\mathcal{F}^1$ to produce $\mathcal{F}^2 = \{\bm{f}^2_i \in \mathbb{R}^{C_2}\}_{i=1}^{rN}$, which is finally used to regress the output point subset $\mathcal{Q} = \{\bm{q}_i \in \mathbb{R}^3\}_{i=1}^{rN}$.

\vspace{-0.1cm}
\subsubsection{Feature Extraction}

We use the feature extractor proposed in \cite{MPU, PUGAN} to learn point-wise features $\mathcal{F}^1 = \{\bm{f}^1_i\}_{i=1}^{N}$, which adopts EdgeConv \cite{wang2019dynamic} as its basic blocks. Specifically, the feature extractor stacks $4$ dense EdgeConv modules via skip-connections. Each EdgeConv acts on local graphs that are dynamically updated based on feature similarities; by max-pooling the embedded features of graph edges, it is advantageous in capturing non-local features of each point. More specifics of the feature extractor are given in the supplementary material.

\vspace{-0.1cm}
\subsubsection{Curvature-Adaptive Feature Expansion}
\label{SecCurvAdptFeaExpand}

This is the key module in our proposed CAD-PU to achieve point set upsampling. It expands at a rate $r$ the input $\mathcal{F}^1 = \{\bm{f}^1_i\}_{i=1}^{N}$ of $N$ point-wise features as $\mathcal{F}^2 = \{\bm{f}^2_i\}_{i=1}^{rN}$ of $rN$ ones. A common practice of feature expansion duplicates $r-1$ copes of $\mathcal{F}^1$ to have $\mathcal{F}^2$, which is adopted in the recent works \cite{PUNet,MPU,PUGAN}. Since each point-wise feature $\bm{f}^1 \in \mathbb{R}^{C_1}$ serves as a function approximator that encodes the geometry centered around a local surface patch, as analyzed in Section \ref{SecMotivation}, this practice is expected to reduce the approximation error w.r.t. the underlying surface $\mathcal{S}^{*}$ by covering $\mathcal{S}^{*}$ with an increased number of surface patch approximators, where the improved covering follows the same distribution as how the input $N$ points distribute on $\mathcal{S}^{*}$. Given a fixed budget of points, Section \ref{SecMotivation} suggests that a better approximation is achieved when the distribution of points on a surface is \emph{curvature-adaptive}, i.e., more points and their corresponding local approximators are located around high-curvature surface areas. However, for the task of point set upsampling, the distribution of input points is \emph{practically not guaranteed} to be curvature-adaptive; consequently, feature expansion used in \cite{PUNet,MPU,PUGAN} is suboptimal in terms of achieving a better approximation of the underlying $\mathcal{S}^{*}$.

In this work, we opt for a curvature-adaptive feature expansion scheme whose specifics are as follows. Given the input $\mathcal{P} = \{\bm{p}_i\}_{i=1}^N$, we first approximate the \emph{mean curvature} $\kappa(\bm{p})$ for each $\bm{p} \in \mathcal{P}$ via computation of the \emph{surface variation} \cite{pauly2002efficient} at $\bm{p}$. Let $\mathcal{N}(\bm{p})$ contain the $k$ nearest neighbors of $\bm{p}$ in $\mathcal{P}$. The covariance matrix $\bm{V}(\bm{p}) \in \mathbb{R}^{3\times 3}$ can be constructed from $\mathcal{N}(\bm{p})$ as $\bm{V}(\bm{p}) = \frac{1}{k} \sum_{\bm{p}' \in \mathcal{N}(\bm{p})} (\bm{p}'-\bm{p})(\bm{p}'-\bm{p})^{\top}$. Let the three eigenvalues of $\bm{V}(\bm{p})$ be $\lambda_3 \geq \lambda_2 \geq \lambda_1 \geq 0$, and the mean curvature $\kappa(\bm{p})$ at $\bm{p}$ is simply approximated as $\lambda_1 / (\lambda_1 + \lambda_2 + \lambda_3)$.
Given $\{\kappa_i\}_{i=1}^N$ corresponding to input points $\{\bm{p}_i\}_{i=1}^N$ and their features $\mathcal{F}^1 = \{\bm{f}^1_i\}_{i=1}^{N}$, our curvature-adaptive feature expansion samples point-wise features from $\mathcal{F}^1$ according to the following probability
\vspace{-0.05cm}
\begin{equation}\label{EqCurvAaptSamplingProb}
w_i = \frac{\log(\kappa_i + 1 + \epsilon)}{\sum_{j=1}^N \log(\kappa_j + 1 + \epsilon)} ,
\vspace{-0.05cm}
\end{equation}
where $\epsilon$ is a small constant and $\sum_{i=1}^N w_i = 1$. We use the logarithmic function in (\ref{EqCurvAaptSamplingProb}) to stabilize computation of $\{w_i\}_{i=1}^N$ in case that the individual curvatures in  $\{\kappa_i\}_{i=1}^N$ vary severely. It is straightforward to know that given (\ref{EqCurvAaptSamplingProb}), point-wise features with high curvatures are sampled more frequently. However, due to the approximate nature of $\{\kappa_i\}_{i=1}^N$ computed via surface variation from the possibly sparse set of $\mathcal{P}$, feature expansion relying purely on probabilistic sampling according to $\{w_i\}_{i=1}^N$ may produce noisy results that deviate from the original shape defined by $\mathcal{P}$, as indicated by the ablation studies in Section \ref{ExpAblation}. We instead use a hybrid manner to practically implement our proposed curvature-adaptive feature expansion. Specifically, given $\mathcal{F}^1$, we \emph{uniformly} sample $\alpha rN$, $0 \leq \alpha \leq 1$, point-wise features and \emph{probabilistically} sample $(1-\alpha)rN$ ones according to $\{w_i\}_{i=1}^N$, and combine them to form $\widehat{\mathcal{F}}^2 = \{ \hat{\bm{f}}_i^2 \in \mathbb{R}^{C_1} \}_{i=1}^{rN}$ of $rN$ features. Such a hybrid manner makes a balance between preserving the geometries defined by $\mathcal{P}$ and adapting distribution to high-curvature areas during expansion of point-wise features, and achieves high-quality upsamplings in practice.

There will be \emph{on average} $r$ duplicate copies of point-wise features in $\widehat{\mathcal{F}}^2$ that are from a same $\bm{f}^1 \in \mathcal{F}^1$. Effective point set upsampling requires injection of discrimination into each group of duplicate features, in order to learn features of higher layer for generation of upsampled points that better approximate the underlying surface $\mathcal{S}^{*}$. To this end, we first track indices of the sampled $rN$ features and organize them into groups each of which contains duplicates of certain $\bm{f}^1 \in \mathcal{F}^1$. We follow the trick used in \cite{PUGAN} by pre-fixing a 2D plane, on which a regular grid of 2D corner points are defined whose coordinates (e.g., $[1, 0]$, $[1, 1]$, $[0, 1]$, $[-1, 1]$, $[-1,0]$, etc) spread away from the origin. Without loss of generality, let $\widehat{\mathcal{G}}^2 = \{\hat{\bm{f}}_i^2 \}_{i=1}^{|\widehat{\mathcal{G}}^2|}$ denote a group of such duplicate features, and $\{ \bm{x}_i \in \mathbb{R}^2 \}_{i=1}^{|\widehat{\mathcal{G}}^2|}$ denote the corresponding set of ordered grid corners on the 2D plane. We augment each $\hat{\bm{f}}_i^2 \in \widehat{\mathcal{G}}^2$ with the corresponding $\bm{x}_i$, giving rise to the new feature $[\hat{\bm{f}}_i^2; \bm{x}_i] \in \mathbb{R}^{C_1 +2}$. The augmentation applies to all groups of duplicate features, followed by a multilayer perceptron (MLP) that learns to map $(C_1 + 2)$-dimensional point-wise features as $C_2$-dimensional ones, and produces the output $\mathcal{F}^2 = \{ \bm{f}_i^2 \in \mathbb{R}^{C_2} \}_{i=1}^{rN}$. Fig. \ref{FigNetwork} gives the illustration. Note that the pre-fixed 2D plane is arbitrarily defined, and the augmented coordinates of grid points on the plane play roles spiritually similar to \emph{anchors} used in object detection \cite{girshick2015fast,ren2015faster}, from which geometric features of upsampled points are learned in subsequent network layers.

\vspace{-0.2cm}
\subsubsection{Regression of Upsampled Points}
\label{sec 3.2.3}

Given the expanded point-wise features in $\mathcal{F}^2 = \{ \bm{f}_i^2\}_{i=1}^{rN}$, we simply use an MLP of 3 fully-connected (FC) layers to regress coordinates of upsampled points that give the final output $\mathcal{Q} = \{ \bm{q}_i \in \mathbb{R}^3 \}_{i=1}^{rN}$ of our proposed CAD-PU. $\mathcal{Q}$ is expected to better approximate the underlying surface $\mathcal{S}^{*}$ by a denser and more curvature-adaptive surface covering. To achieve the goal, we propose loss functions that instantiate the upsampling objective motivated in Section \ref{SecMotivation} to train CAD-PU, as presented shortly.

\vspace{-0.1cm}
\subsection{Training Objectives}
\label{SecTrainingObj}

Given the training set $\{\mathcal{P}_i, \mathcal{Q}_i^{*}\}_{i=1}^M$, analysis in Section \ref{SecMotivation} motivates a learning criterion of surface curvature minimization. Using the motivated criterion as a regularizer $R(\cdot)$ gives the following objective to train the generator $G$ of CAD-PU
\vspace{-0.1cm}
\begin{equation}\label{EqnMainObj}
J(G; \{\mathcal{P}_i, \mathcal{Q}_i^{*}\}_{i=1}^M) = \sum_{i=1}^M L(G(\mathcal{P}_i), \mathcal{Q}_i^{*}) + \beta\cdot R(G(\mathcal{P}_i)) ,
\vspace{-0.15cm}
\end{equation}	
where $\beta$ is a penalty parameter, and $L(\cdot,\cdot)$ is a distance of point set whose use is to make the resulting $\mathcal{Q} = G(\mathcal{P})$ as close to the training ground-truth $Q^{*}$ as possible. The use of regularization $R(G)$ is important to make the learned $G$ generalize to testing samples. We specify our realizations of $L(\cdot,\cdot)$ and $R(\cdot)$ as follows.

Given any training pair $\{\mathcal{P}, \mathcal{Q}^{*}\}$ and the prediction $\mathcal{Q} = G(\mathcal{P})$, we follow \cite{fan2017point} and use Earth Mover's Distance (EMD) as a distance measure between point sets
\vspace{-0.1cm}
\begin{equation}\label{EqnEMD}
L(\mathcal{Q}, \mathcal{Q}^{*}) = \min_{\psi : \mathcal{Q} \rightarrow \mathcal{Q}^{*} } \sum_{ \bm{q} \in \mathcal{Q} } \| \bm{q} - \psi(\bm{q}) \|_2 ,
\vspace{-0.1cm}
\end{equation}
where $\psi : \mathcal{Q} \rightarrow \mathcal{Q}^{*}$ indicates a bijection. EMD in fact solves an assignment problem by finding the optimal one-to-one correspondence between points in $\mathcal{Q}$ and $\mathcal{Q}^{*}$, such that the sum of their point-wise distances is the smallest. To efficiently compute EMD, we use the $(1+\epsilon)$ approximation scheme \cite{Bertsekas85}.

The regularizer $R(G)$ is motivated to learn a generator $G(\cdot)$ such that the resulting $\mathcal{Q} = G(\mathcal{P})$ for any input $\mathcal{P}$ represents a surface $\mathcal{S}^{*}$ whose local curvatures are small; consequently, the approximation error $E(\mathcal{S}_{\mathcal{Q}}, \mathcal{S}^{*})$ in (\ref{EqnErrBoundQ}) is bounded to be small as well. A surface with smaller values of local curvatures also satisfies the desired surface property of smoothness and fairness \cite{botsch2010polygon}. To implement the regularizer, given the discrete point set of prediction $\mathcal{Q}$, we rely on the following surrogate that approximately computes the discrete version of \emph{mean curvature} centered at any $\bm{q} \in \mathcal{Q}$. This is similar to the computation of mean curvature via surface variation in Section \ref{SecCurvAdptFeaExpand}, but we here incorporate the generator $G$, i.e., $\mathcal{Q} = G(\mathcal{P})$, into the computation such that parameters of $G$ can be optimized
\vspace{-0.1cm}
\begin{equation}\label{EqnCurvaturebyEst}
\kappa(\bm{q}; \mathcal{Q}) =  \frac{1}{k} \sum_{ \bm{q}' \in \mathcal{N}(\bm{q})} \left| \left< (\bm{q}' - \bm{q}) / \| \bm{q}' - \bm{q} \|_2, \bm{n}_{\bm{q}} \right> \right| ,
\vspace{-0.1cm}
\end{equation}
where $\mathcal{N}(\bm{q})$ contains the $k$ nearest neighbors of $\bm{q}$ in $\mathcal{Q}$, and $\bm{n}_{\bm{q}} \in \mathbb{R}^3$ denotes the unit normal vector of the surface at $\bm{q}$. The term (\ref{EqnCurvaturebyEst}) measures the averaged angle between the normal vector and the vector defined by pointing $\bm{q}$ towards each $\bm{q}'$ of its neighboring points.
Since $\bm{n}_{\bm{q}}$ is orthogonal to the tangent plane of the surface at $\bm{q}$, each inner product in (\ref{EqnCurvaturebyEst}) characterizes how the normals vary directionally in the local neighborhood $\mathcal{N}_{\bm{q}}$, thus approximately measuring the local, directional curvature, and an average of $| \mathcal{N}_{\bm{q}} | = k$ inner products in (\ref{EqnCurvaturebyEst}) approximately measures the local, mean curvature. While $\bm{n}_{\bm{q}}$ can be approximated via eigen-decomposition of the covariance matrix $\bm{V}(\bm{q})$ constructed from $\mathcal{N}(\bm{q})$, in practice, we use $\bm{n}_{\bm{q}^{*}}$ associated with the point in the ground-truth $\mathcal{Q}^{*}$ that is closest to $\bm{q}$, which can be pre-computed and efficiently retrieved during the network training. Except for $\kappa(\bm{q}; \mathcal{Q})$ of (\ref{EqnCurvaturebyEst}), we want to further leverage the local geometries in the ground-truth $\mathcal{Q}^{*}$ to regularize the learning of $\mathcal{Q} = G(\mathcal{P})$, and compute the following discrete notion that bears similarity with mean curvature for a local surface neighborhood $\mathcal{N}^{*}(\bm{q})$ defined by embedding any $\bm{q} \in \mathcal{Q}$ into its ground-truth $\mathcal{Q}^{*}$
\begin{equation}\label{EqnCurvaturebyGT}
\widetilde{\kappa}(\bm{q}; \mathcal{Q}^{*}) =  \frac{1}{k} \sum_{ {\bm{q}^{*}}' \in \mathcal{N}^{*}(\bm{q})} \left| \left< ({\bm{q}^{*}}' - \bm{q}) / \| {\bm{q}^{*}}' - \bm{q} \|_2, \bm{n}_{{\bm{q}^{*}}'} \right> \right| ,
\end{equation}
where $\mathcal{N}^{*}(\bm{q})$ contains the $k$ nearest neighbors of $\bm{q}$ in $\mathcal{Q}^{*}$, and $\bm{n}_{{\bm{q}^{*}}'}$ denotes the unit normal vector at the neighbor ${\bm{q}^{*}}'$. Combining (\ref{EqnCurvaturebyEst}) and (\ref{EqnCurvaturebyGT}) gives our realization of the regularizer $R(\mathcal{Q}) = R(G(\mathcal{P}))$
\begin{equation}\label{EqnRegu}
R(\mathcal{Q}; \mathcal{Q}^{*}) = \frac{1}{rN} \sum_{\bm{q} \in \mathcal{Q}} \kappa(\bm{q}; \mathcal{Q}) + \widetilde{\kappa}(\bm{q}; \mathcal{Q}^{*}) .
\end{equation}
Note that (\ref{EqnRegu}) averages over $rN$ discrete notions of absolute values corresponding to the $rN$ points of $\{ \bm{q}_i \in \mathcal{Q} \}_{i=1}^{rN}$; minimizing (\ref{EqnRegu}) is thus robust to allow a small portion of them to have relatively higher curvature values, while achieving a smaller value of total curvature.

\vspace{0.1cm}
\noindent\textbf{A Strategy of Adversarial Training} Adversarial training has shown its power in learning generative models, including those for generation of point sets \cite{achlioptas2018learning,PUGAN}. Following \cite{PUGAN}, we introduce a discriminator $D(\cdot)$ that is trained to differentiate the distribution of $\{ \mathcal{Q} = G(\mathcal{P}) \}$ from that of $\{ \mathcal{Q}^{*} \}$, giving rise to the following problem
\vspace{-0.15cm}
\begin{eqnarray}\label{EqnAdvTrainObj}
&& \min_{G}  J(G; \{\mathcal{P}_i, \mathcal{Q}_i^{*}\}_{i=1}^M) +  \frac{\gamma}{2} \sum_{i=1}^M [ D(G(\mathcal{P}_i)) -1 ]^2 , \\
&& \min_{D} \frac{1}{2} \sum_{i=1}^M D(G(\mathcal{P}_i))^2 + [ D(\mathcal{Q}_i^{*}) - 1 ]^2 ,
\end{eqnarray}
where $\gamma$ is a weight parameter. We use PointNet \cite{qi2017pointnet} to implement our discriminator $D(\cdot)$, which serves as a simple and effective model of point set classification.

\vspace{-0.1cm}
\section{Experiments}\label{SecExperiments}
\subsection{Setups}
\label{ExpSetups}
\vspace{-0.05cm}
We use a benchmark dataset of 147 object surface models collected by \cite{PUGAN} for our experiments. These models include simple and smooth ones (e.g., a cup) and also geometrically complex ones (e.g., a statue of Buddha). Following \cite{PUGAN}, we use the same 120 objects in the dataset as training models and use the remaining ones for testing.  For each training object, we follow \cite{PUGAN} and segment its surface into 200 overlapped patches, resulting in a total of $24,000$ training surface patches; training of our CAD-PU and other deep learning methods \cite{PUNet,MPU,PUGAN} are conducted on the surface patch level. We prepare the input and output pairs of training point sets as follows. For each surface patch, we uniformly sample $256$ points as training input; for training output, we first uniformly sample $10,000$ points from each surface patch and compute their point-wise curvatures (cf. Section \ref{SecCurvAdptFeaExpand}), and we then select from them $1,024$ points in a curvature-adaptive manner based on point-wise probabilities computed by (\ref{EqCurvAaptSamplingProb}). Note that for comparison with existing methods \cite{PUNet,MPU,PUGAN} whose training outputs are uniformly sampled from object surfaces, we simply uniformly sample $1,024$ points from each surface patch as their training output. In Section \ref{SecExperimentSOTA}, we also investigate how these methods perform when using curvature-adaptive point sets as their training outputs.

We conduct testing of different methods as follows. For each testing object, we uniformly sample $2,048$ points from its surface and cluster them into $8$ subsets of equal size based on spatial proximity; the obtained $256$ points per subset are used as input of patch-wise upsampling. Given the default upsampling rate of $r=4$, we produce $1,024$ output points for each subset, and the final result is formed by a simple aggregation of $1,024\times 8 = 8,192$ points. Given the different choices of training output point sets used in our CAD-PU and existing methods \cite{PUNet,MPU,PUGAN}, for a meaningful comparison, we choose to evaluate the upsampling result by measuring how better it can reconstruct the underlying surface. Specifically, for an upsampled result of $8,192$ points, we first use the de-facto standard method of screened Poisson reconstruction \cite{kazhdan2013screened} to reconstruct its surface mesh, and then uniformly sample two sets of $100,000$ points respectively from the reconstructed surface and the ground-truth one, which enables us to measure the error of surface reconstruction based on point set distance. In this work, we use Chamfer distance (CD) and Hausdorff distance (HD) as the metrics of point set distance. We note that such a quantitative evaluation is more consistent with visual perception of point set upsampling as an improved approximation of the underlying surface.

Implementation details of our CAD-PU are as follows. We use Adam \cite{Adam} to train the generator and discriminator under a two time-scale update rule (TTUR) \cite{NIPS2017_7240} for $120$ epochs, where the initial learning rates are $0.001$ and $0.0001$, respectively. Learning rates are decayed by $0.7$ every $50,000$ iterations, and the batch size is $28$. We set the hyperparameters $\alpha$, $\beta$, $\gamma$, $k$, and $\epsilon$ respectively as $0.5$, $0.15$, $0.005$, $12$, and $0.01$, which work stably well.

\vspace{-0.2cm}
\subsection{Comparisons with Existing Methods}
\label{SecExperimentSOTA}

We compare our CAD-PU with the state-of-the-art deep learning methods of PU-Net \cite{PUNet}, MPU \cite{MPU}, and PU-GAN \cite{PUGAN}, and also with the optimization-based EAR \cite{huang2013edge}. Results of these methods are obtained using their publicly released codes, with tuning of their optimal hyperparameters. The methods of PU-Net, MPU, and PU-GAN are trained by output point sets uniformly sampled from object surface patches, which are different from ours; except for direct comparisons, it is also interesting to observe how these methods perform when using the same training outputs of curvature-adaptive point sets as our CAD-PU does.

\begin{figure*}[!t]
\centering
\vspace{-0.2cm}
\includegraphics[width=0.98\textwidth]{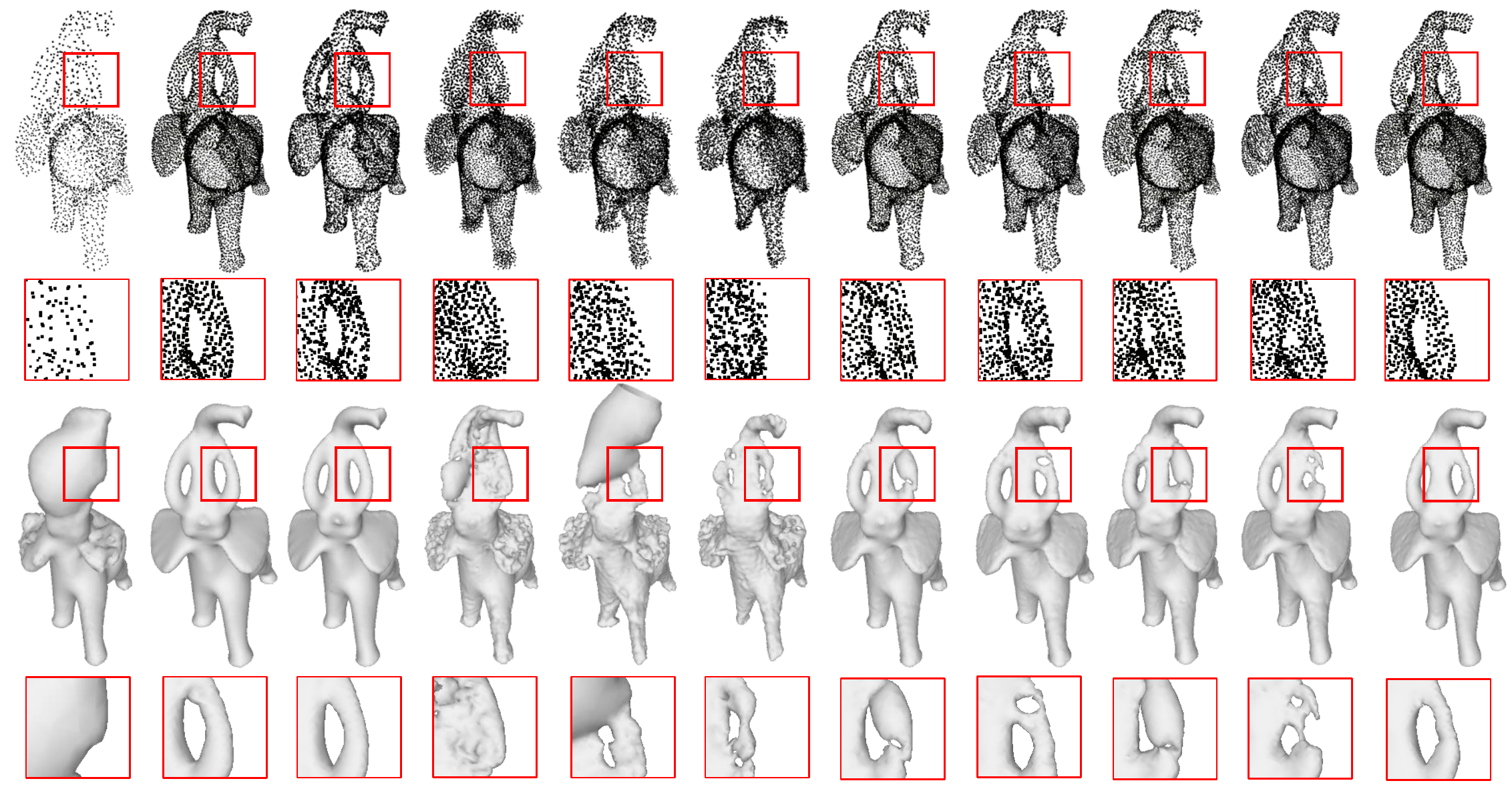}

\vspace{0.1cm}
\footnotesize{
\begin{tabular}{p{1.2cm}<{\centering} p{1.2cm}<{\centering} p{1.2cm}<{\centering} p{1.2cm}<{\centering} p{1.2cm}<{\centering} p{1.2cm}<{\centering} p{1.2cm}<{\centering} p{1.2cm}<{\centering} p{1.2cm}<{\centering} p{1.2cm}<{\centering} p{1.2cm}<{\centering}}
Input & GT & GT & EAR & PU-Net & PU-Net & MPU & MPU & PU-GAN & PU-GAN  & CAD-PU\\
 & (uni.) & (cur.) & & (uni.) & (cur.) & (uni.) & (cur.) & (uni.) & (cur.) & \\
\end{tabular}}
\vspace{-0.12cm}
\caption{Qualitative comparisons of different methods on an testing object of elephant. PU-Net (uni.) and PU-Net (cur.) mean that the results are obtained by training respectively with \emph{uniformly} and \emph{curvature-adaptively} distributed point sets. The same applies to MPU and PU-GAN. \emph{More comparative results on other testing objects are shown in the supplementary material.} }
\vspace{-0.1cm}
\label{FigSOTA}
\end{figure*}

Quantitative results in Table \ref{TableSOTA} show that our CAD-PU outperforms all existing methods, no matter what training point sets they use; training MPU and PU-GAN using curvature-adaptive point sets may not bring benefits, suggesting that network designs and learning objectives of these methods are suboptimal in improving surface approximation via point set upsampling. Qualitative results of different methods on an example of elephant are shown in Fig. \ref{FigSOTA}, where each upsampling result is accompanied with its surface reconstruction. Superiority of our method over existing ones is consistent with those observed in Table \ref{TableSOTA}. Our CAD-PU is particularly advantageous in recovering complex surface geometries at high-curvature areas, e.g.,  thin structures of surface, while other methods may wrongly glue together spatially close, but originally isolated surface parts. More qualitative results of other testing objects are shown in the supplementary material.

\begin{table}[!t]
	\caption{Quantitative comparisons among different methods. Results are obtained by averaging over the $27$ testing objects of the dataset collected in \cite{PUGAN}. We evaluate point set upsampling results based on how they can reconstruct the underlying surfaces. We use Chamfer distance (CD) and Hausdorff distance (HD) as the point set based metrics of surface reconstruction error. Refer to Section \ref{ExpSetups} for how the errors are computed.}
	\label{TableSOTA}
	\centering
    \begin{tabular}{c|cc}
	\hline
	Methods & CD($10^{-3}$) & HD($10^{-3}$) \\
	\hline
	EAR\cite{huang2013edge} & $1.44$ & $20.75$ \\
	\hline
	PU-Net \cite{PUNet} & $2.25$ & $30.53$ \\
	PU-Net \cite{PUNet} with curvature-adaptive GT & $1.57$ & $21.76$ \\
	\hline
	MPU \cite{MPU} & $0.85$ & $19.24$ \\
	MPU \cite{MPU} with curvature-adaptive GT & $2.38$ & $36.87$ \\
	\hline
	PU-GAN \cite{PUGAN} & $0.72$ & $16.07$ \\
	PU-GAN \cite{PUGAN} with curvature-adaptive GT & $1.55$ & $25.14$ \\
	\hline
	CAD-PU & $\mathbf{0.57}$ & $\mathbf{15.83}$\\
	\hline
    \end{tabular}
    \vspace{-0.25cm}
\end{table}

\subsection{Ablation Studies}
\label{ExpAblation}
\vspace{-0.05cm}

\begin{figure*}[!t]
\centering
\vspace{-0.4cm}
\includegraphics[width=0.9\textwidth]{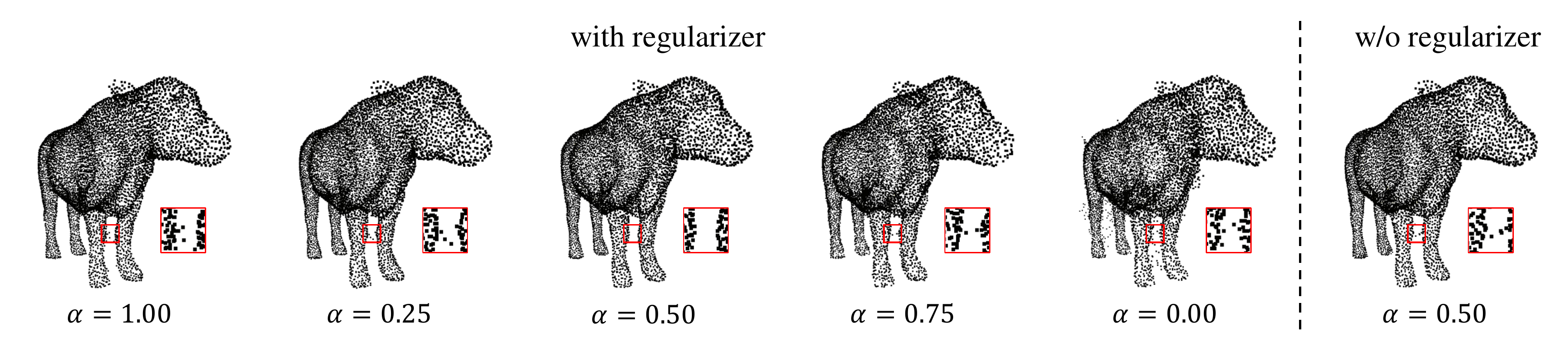}
\vspace{-0.4cm}
\caption{Qualitative results of ablation studies on the effects of curvature-adaptive feature expansion and the regularizer \eqref{EqnRegu} in our proposed CAD-PU, where the curvature-adaptive degree is controlled by the values of $\alpha$. }
\label{FigAblation}
\vspace{-0.2cm}
\end{figure*}

In this section, we conduct ablation studies on the following two aspects of CAD-PU. Table \ref{TableAblationStudy} reports the quantitative results, with qualitative example results shown in Fig. \ref{FigAblation}.

\noindent \textbf{Effect of curvature-adaptive feature expansion} We first evaluate the effect of curvature-adaptive feature expansion in our CAD-PU, by controlling the values of $\alpha$. When $\alpha=1$, all the features of input points are uniformly sampled for expansion by $r$ times; when $\alpha=0$, they are sampled proportionally based on their curvatures. Table \ref{TableAblationStudy} shows that the best performance achieves at $\alpha=0.5$, confirming the advantage of our hybrid sampling scheme for point set upsampling via feature expansion.

\noindent \textbf{Effect of the regularizer \eqref{EqnRegu}} The regularizer \eqref{EqnRegu} is proposed to improve the surface approximation via point set upsampling. Its efficacy is verified in Table \ref{TableAblationStudy} in terms of both CD and HD metrics. In terms of visual quality, the regularizer is particularly helpful to suppress generation of outlier points, as shown by the example in Fig. \ref{FigAblation}.

\begin{table}[!t]
	\caption{Ablation studies on our proposed CAD-PU. We report averaged results of Chamfer distance (CD) and Hausdorff distance (HD) over the $27$ testing objects of the dataset collected in \cite{PUGAN}.}
	\vspace{-0.2cm}
	\label{TableAblationStudy}
	\centering
	\begin{tabular}{c|c|cc}
		\hline
		$\alpha $ & regularizer \eqref{EqnRegu}  & CD$(10^{-3})$ & HD$(10^{-3})$ \\
		\hline
		$0.50$ & $\times$ & $2.16$ & $40.54$\\
		\hline
		$1.00$ & \checkmark & $1.81$ & $33.18$\\
		$0.75$ & \checkmark & $1.26$ & $21.90$ \\
		$0.50$ & \checkmark & $\mathbf{0.57}$ & $\mathbf{15.83}$\\
		$0.25$ & \checkmark & $1.83$ & $30.86$\\
		$0.00$ & \checkmark & $2.58$ & $51.72$ \\
		\hline
	\end{tabular}
	\vspace{-0.1cm}
\end{table}

\subsection{Testing of Robustness}
\vspace{-0.05cm}
We evaluate the robustness of our method in terms of the following two aspects, by comparing with the state-of-the-art method of PU-GAN \cite{PUGAN}.

\begin{figure*}[!t]
\centering
\includegraphics[width=0.9\textwidth]{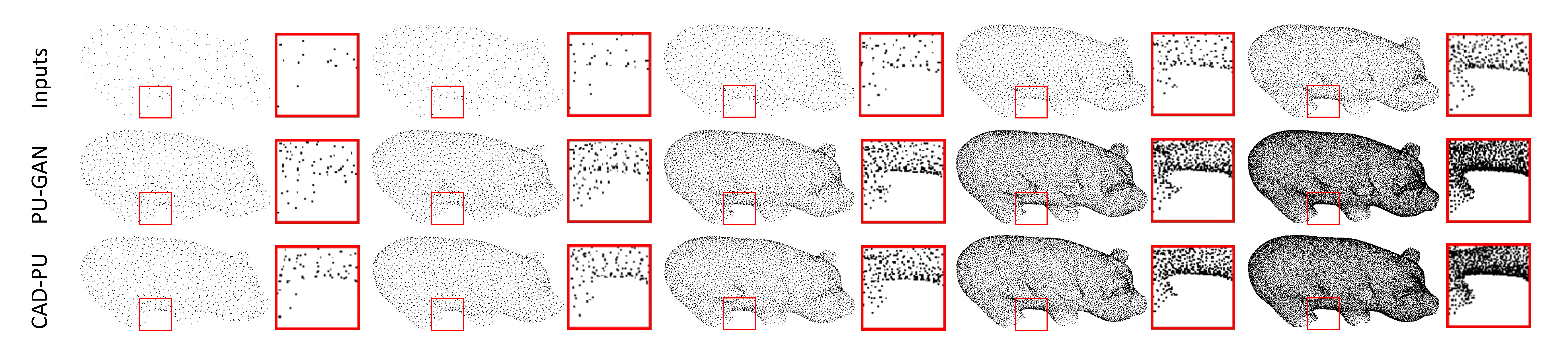}
\vspace{-0.3cm}
\caption{Qualitative results of robustness testing against varying working scales of spatial extent. The numbers of total input points per testing object (from left to right) are $256$, $512$, $1,024$, $2,048$, and $4,096$, respectively.}
\vspace{-0.3cm}
\label{FigSize}
\end{figure*}

\begin{table}[!t]
 \caption{Testing of robustness against varying scales of spatial extent. We report averaged results of Chamfer distance (CD $\times 10^{-3}$) and Hausdorff distance (HD $\times 10^{-3}$) over the $27$ testing objects of the dataset collected in \cite{PUGAN}. }
 \label{TableExpSize}
 \centering
 \vspace{-0.2cm}
 \resizebox{0.49\textwidth}{!}{
 \begin{tabular}{c|c|ccccc}
  \hline
  \multirow{2}{*}{Metrics} & \multirow{2}{*}{Methods} & \multicolumn{5}{c}{Number of input points per object}\\ \cline{3-7}
  & & $256$ &  $512$ & $1,024$ & $2,048$ & $4,096$ \\
  \hline
        \multirow{2}{*}{CD} & PU-GAN\cite{PUGAN} & $2.63$ & $2.13$ & $1.29$ & $0.72$ & $0.67$\\
         & CAD-PU & $\mathbf{2.11}$ & $\mathbf{1.64}$ & $\mathbf{1.16}$ & $\mathbf{0.57}$ & $\mathbf{0.37}$ \\

        \hline
        \multirow{2}{*}{HD} & PU-GAN\cite{PUGAN}  & $40.99$ & $32.90$ &  $21.09$ &  $16.07$ & $15.56$\\
        & CAD-PU & $\mathbf{38.61}$ & $\mathbf{29.68}$ & $\mathbf{19.41}$ & $\mathbf{15.83}$ & $\mathbf{11.52}$\\
  \hline
 \end{tabular}}
 \vspace{-0.1cm}
\end{table}

\begin{table}[!t]
 \caption{Testing of robustness against varying levels of input noise. We add point-wise Gaussian perturbations with different standard deviations to input points of the $27$ testing objects collected in \cite{PUGAN}. Averaged results of Chamfer distance (CD $\times 10^{-3}$) and Hausdorff distance (HD $\times 10^{-3}$) are reported. }
 \label{TableExpNoise}
 \centering
 \vspace{-0.2cm}
 \resizebox{0.49\textwidth}{!}{
 \begin{tabular}{c|c|ccccc}
  \hline
  \multirow{2}{*}{Metrics} & \multirow{2}{*}{Methods} & \multicolumn{5}{c}{Standard deviations of input noise}\\ \cline{3-7}
  & & $0$ &  $0.001$ & $0.005$ & $0.01$ & $0.02$ \\
  \hline
        \multirow{2}{*}{CD} & PU-GAN\cite{PUGAN} & $0.72$ & $0.86$ & $1.14$ & $1.54$ & $2.62$\\
         & CAD-PU & $\mathbf{0.57}$ & $\mathbf{0.78}$ & $\mathbf{1.05}$ & $\mathbf{1.44}$ & $\mathbf{1.87}$ \\
        \hline
        \multirow{2}{*}{HD} & PU-GAN\cite{PUGAN} & $16.07$ & $19.58$ & $24.13$ & $30.16$ & $40.52$\\
        & CAD-PU & $\mathbf{15.83}$ & $\mathbf{18.63}$ & $\mathbf{22.83}$ & $\mathbf{25.03}$ & $\mathbf{33.29}$\\
  \hline
 \end{tabular}}
 \vspace{-0.3cm}
\end{table}

\noindent \textbf{Robustness against varying scales of spatial extent}
Both our method and PU-GAN conduct patch-level upsampling, which upsamples a fixed number of input points per surface patch by a specified factor $r$. When the number of total input points for a testing object varies, it amounts to applying the methods to a varying working scale of spatial extent that contains the same number of points, or equivalently, it amounts to conducting parallel upsamplings at a varying number of surface patches for the testing object. To investigate how our method and PU-GAN perform when the number of input points per testing object varies, we respectively use $256$, $512$, $1,024$, $2,048$, and $4,096$ input points for any testing object, which correspond to conduct parallel upsamplings respectively of $1$, $2$, $4$, $8$, and $16$ surface patches. Fewer working patches suggest that the methods are to recover larger-scale surface geometries via point set upsampling; conversely, more working patches suggest that the methods are to recover smaller-scale details of surface geometries. Results in Table \ref{TableExpSize} tell that across a range of working scales, our CAD-PU outperforms PU-GAN consistently. Qualitative results in Fig. \ref{FigSize} seem suggest that the advantage of our method is more obvious in the regime of smaller working scales.

\begin{figure}[!t]
\centering
\vspace{-0.2cm}
\includegraphics[width=0.45\textwidth]{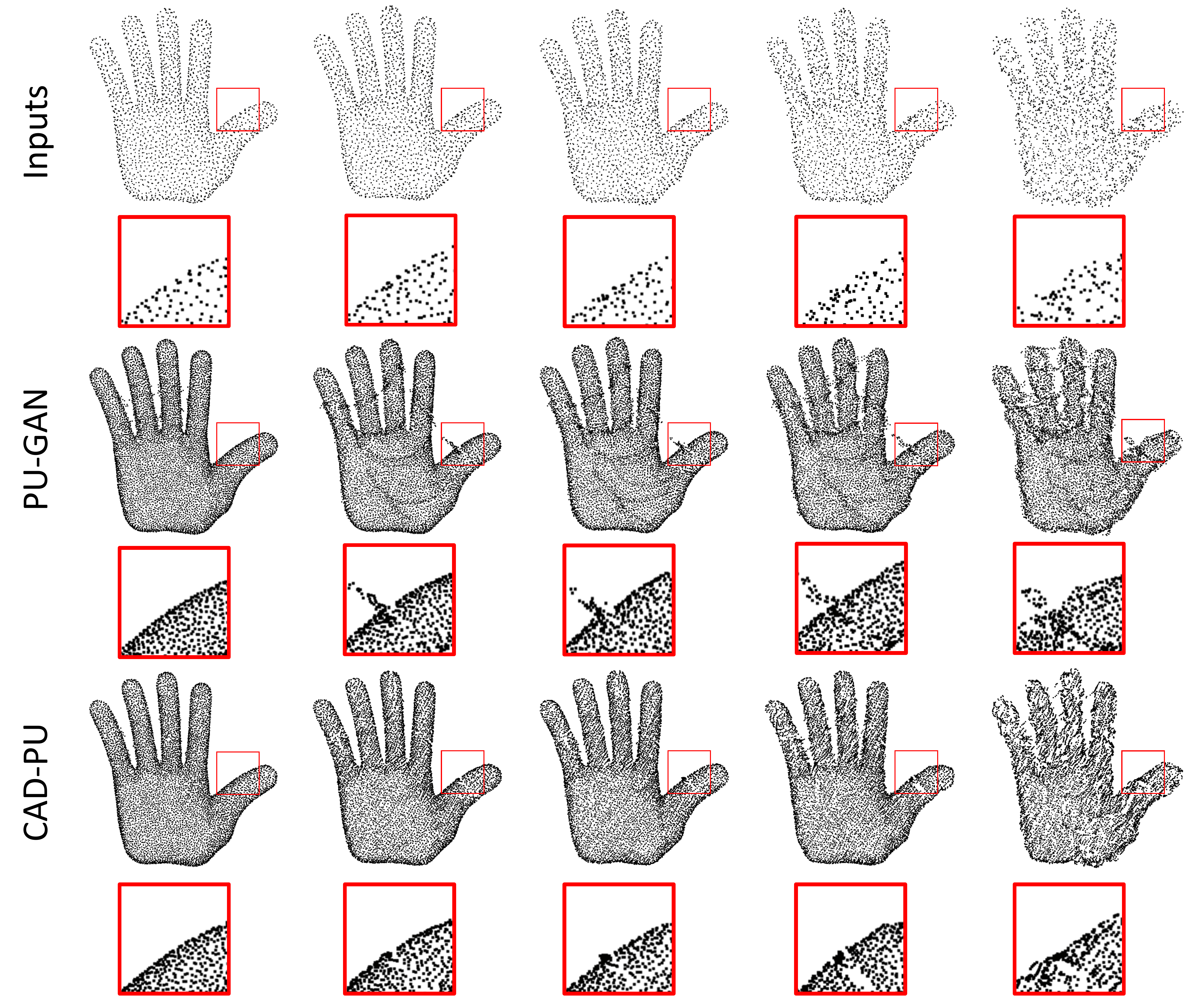}
\vspace{-0.2cm}
\caption{Qualitative results of robustness testing against varying levels of input noise. The standard deviations of point-wise Gaussian perturbations (from left to right) are $0$, $0.001$, $0.005$, $0.01$, and $0.02$, respectively.}
\vspace{-0.1cm}
\label{FigNoise}
\end{figure}

\noindent \textbf{Robustness against varying levels of input noise}
Points obtained in real-world settings inevitably contain noise. To investigate how our method and PU-GAN perform against noise perturbations, we add Gaussian noise of varying levels (with standard deviations of $0.001$, $0.005$, $0.01$, and $0.02$) in a point-wise manner to the testing inputs. Quantitative results in Table \ref{TableExpNoise} show that our method outperforms PU-GAN, with larger margins for noise-contaminated inputs. Examples in Fig. \ref{FigNoise} tell that given noise contaminations, PU-GAN tends to degrade its performance with generation of point outliers, while our results are relatively stable.

\vspace{-0.2cm}
\subsection{Real-world Evaluation}

\begin{figure}[!t]
\centering
\includegraphics[width=0.45\textwidth]{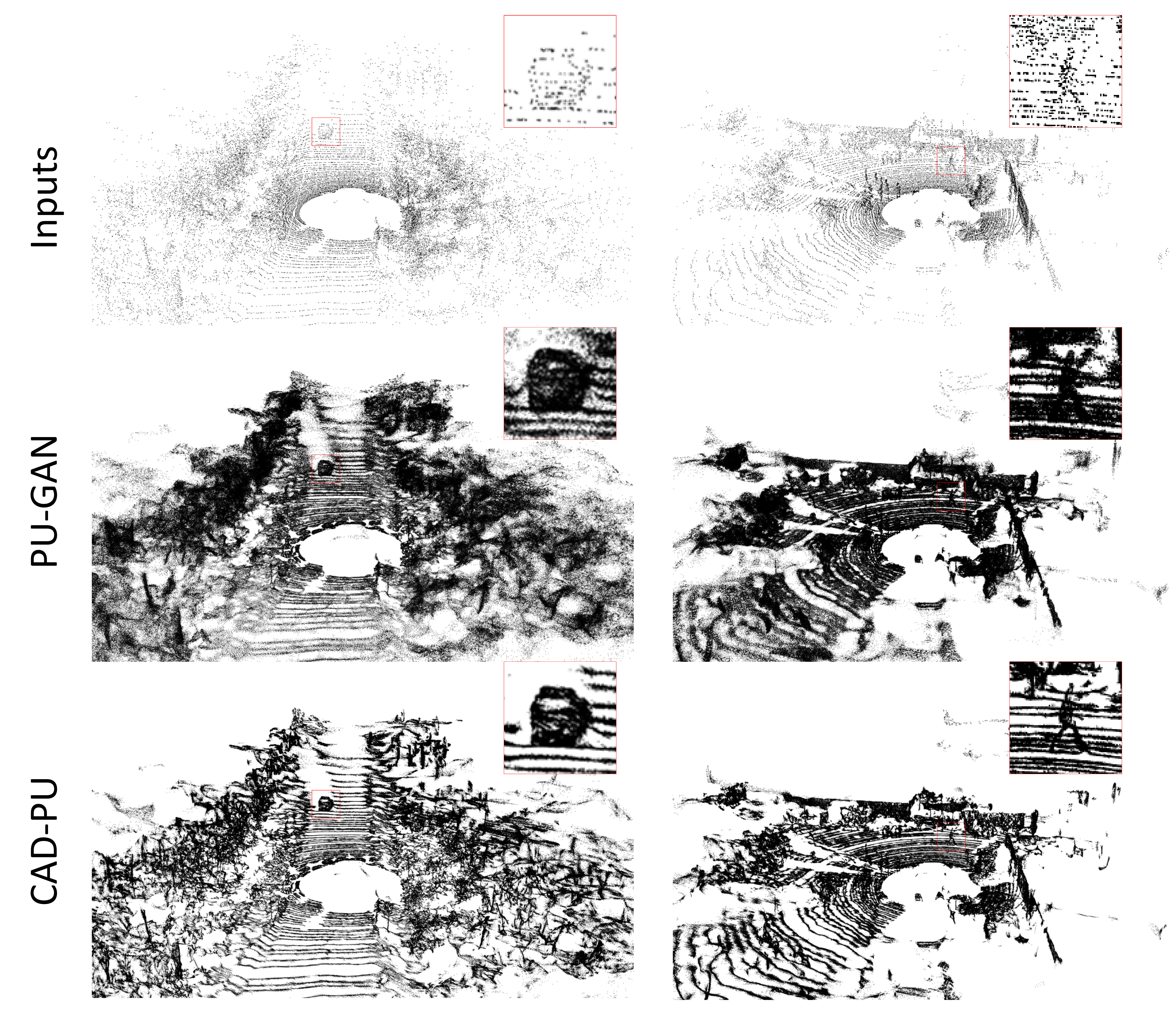}
\vspace{-0.2cm}
\caption{Real-world testing by applying PU-GAN \cite{PUGAN} and our CAD-PU to LiDAR-scanned street scenes from Kitti \cite{Kitti}. }
\vspace{-0.3cm}
\label{FigScan}
\end{figure}

We finally evaluate our method for upsampling point sets obtained from real-world scans. We apply CAD-PU and PU-GAN to LiDAR-scanned street scenes from Kitti \cite{Kitti}, as shown in Fig. \ref{FigScan}. Such real-world settings are challenging since the input points are rather sparse and irregular. Since no ground truth is available, we visualize the comparative results in Fig. \ref{FigScan}. PU-GAN tends to generate points in blank areas, resulting in obscure object boundaries. In contrast, our method recovers sharper geometries at surface boundaries, due to our internal mechanism of high-curvature attention.

\vspace{-0.2cm}
\section{Conclusion}\label{SecConclusion}

This work is motivated to improve surface approximation by learning point set upsamping. To this end, we first analyze the approximation error bounds of the input and output point sets, and identify point-wise curvatures as an important controlling factor that determines the quality of upsampled results. Based on the analysis, we have proposed a novel network design of CAD-PU and the corresponding learning objective. Both quantitative and qualitative results show the advantages of our method over existing ones. Encouraging results are also obtained by applying CAD-PU to LiDAR-scanned street scenes. In future research, we are interested in learning better upsamplings of point sets obtained from real scans of objects and/or scenes, such that downstream tasks of point set analysis would be facilitated.

\ifCLASSOPTIONcaptionsoff
  \newpage
\fi

\vspace{-0.2cm}
\bibliographystyle{IEEEtran}
\bibliography{reference}

\newpage

\section*{Appendix A}\label{appendix:a}
\section*{Network Specifics of CAD-PU}

We have introduced our proposed method of Curvature-ADaptive Point set Upsampling network (CAD-PU) in Sec. 4. In this section, we present specifics of the three network modules.

\begin{figure*}[!b]
\centering
\vspace{0.8cm}
\includegraphics[width=1.0\textwidth]{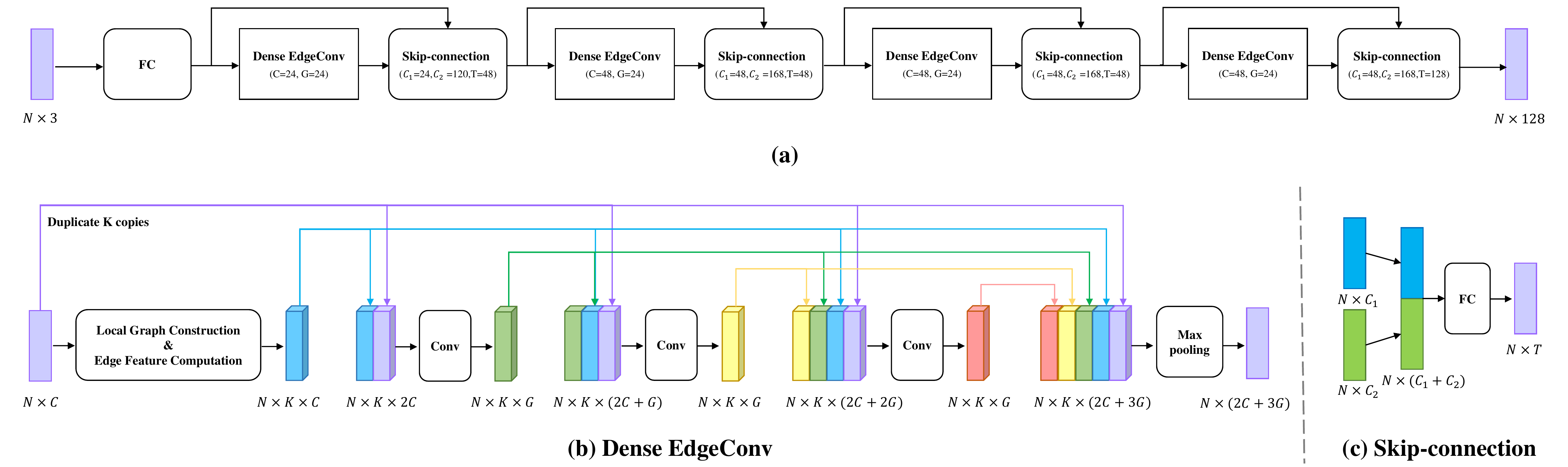}
\vspace{0.1cm}
\caption{(a) An illustration of the feature extractor used in CAD-PU, which is the same as that in \cite{MPU}. It stacks $4$ modules of Dense EdgeConv (b) via skip-connections (c). $N$ is the number of input points, and $C$, $G$, $C_1$, $C_2$, and $T$ denote the respective numbers of feature dimensions used in (b) and (c). In (b), we search $K = 16$ nearest neighbors for each point to define the local graph and compute the corresponding edge-wise features. Color-coded columns represent feature tensors of different sizes. }
\vspace{0.4cm}
\label{FeaExtractIllus}
\end{figure*}

\noindent\textbf{Feature Extraction}
For point-wise feature extraction, we adopt a same extractor as in \cite{MPU, PUGAN}, which stacks $4$ modules of Dense EdgeConv \cite{wang2019dynamic} via skip-connections, as illustrated in Fig. \ref{FeaExtractIllus}-(a). We present its specifics for completion of the present paper. Specifically, the input of each Dense EdgeConv module (cf. Fig. \ref{FeaExtractIllus}-(b)) is the output point-wise features of its previous one, except the first one that takes $24$-dimensional point-wise features lifted from point coordinates by a fully-connected (FC) layer; the local graph for each point in the module is defined by searching its $K$ nearest neighbors based on $l_2$ distances in the feature space, from which $K$ edge features associated with the center point are computed and concatenated with the duplicated, input point-wise feature, followed by 3 densely-connected layers of $1\times 1$ convolution to produce the embedded features; these features are finally max-pooled to produce the output of the Dense EdgeConv module. Each skip-connection takes as input the input and output features of its previous Dense EdgeConv module, and processes them via concatenation and FC based feature transformation, as illustrated in Fig. \ref{FeaExtractIllus}-(c).

\noindent\textbf{Curvature-Adaptive Feature Expansion}
This is the key module in our proposed CAD-PU. We have spelled out its designing details in Sec. 4.1.2. The only parametric component of the module is its final multilayer perceptron (MLP), which is formed by 2 FC layers respectively of $128$ neurons.

\noindent\textbf{Regression of Upsampled Points}
 We use an MLP of 3 FC layers to regress the coordinates of upsampled points. The 3 FC layers have $128$, $64$, and $3$ output neurons, respectively.

\section*{Appendix B}\label{appendix:b}
\section*{Additional Results of Qualitative Comparisons with Existing Methods}

\begin{figure*}[!t]
\centering
\vspace{-0.6cm}
\includegraphics[width=0.90\textwidth]{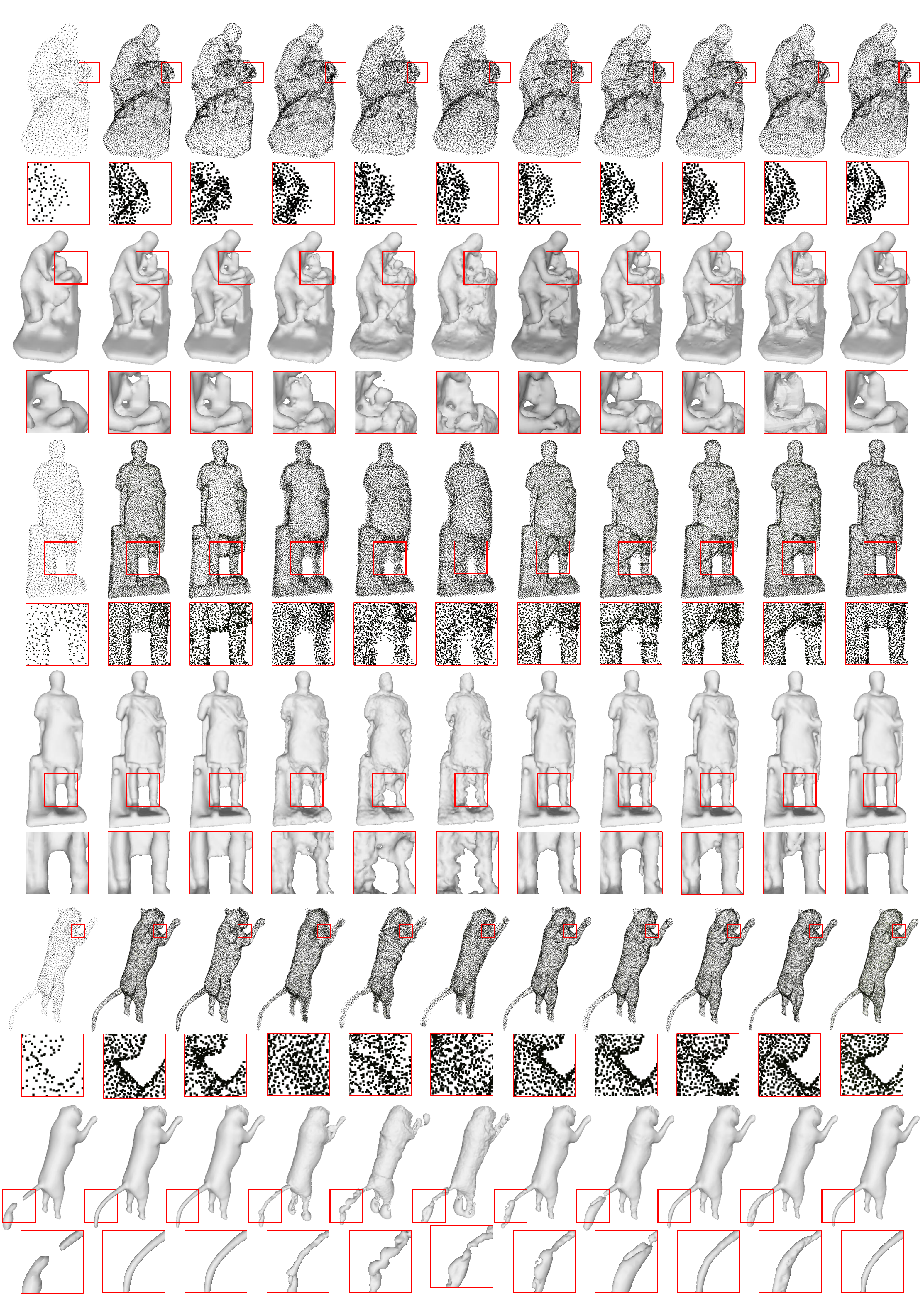}
\footnotesize{
\begin{tabular}{p{0.7cm}<{\centering} p{1.0cm}<{\centering} p{1.0cm}<{\centering} p{0.9cm}<{\centering} p{1.1cm}<{\centering} p{1.0cm}<{\centering} p{1.0cm}<{\centering} p{1.0cm}<{\centering} p{1.1cm}<{\centering} p{1.2cm}<{\centering}p{1.2cm}<{\centering}p{1.2cm}<{\centering} p{1.0cm}<{\centering}}
& Input & GT & GT & EAR & PU-Net & PU-Net & MPU & MPU & PU-GAN & PU-GAN  & CAD-PU &\\
&  & (uni.) & (cur.) & & (uni.) & (cur.) & (uni.) & (cur.) & (uni.) & (cur.) & & \\
\end{tabular}}
\vspace{-0.1cm}
\caption{Qualitative comparisons of different methods on testing objects collected in \cite{PUGAN}. PU-Net (uni.) and PU-Net (cur.) mean that the results are obtained by training respectively with \emph{uniformly} and \emph{curvature-adaptively} distributed point sets. The same applies to MPU and PU-GAN. }
\label{FigSuppSOTA}
\end{figure*}

In Fig. \ref{FigSuppSOTA}, we show more examples of qualitative comparisons among different methods on the testing objects of the dataset collected in \cite{PUGAN}.

\section*{Appendix C}\label{appendix:c}

In Sec. 5.2, we have compared our CAD-PU with existing methods, including the deep learning-based PU-Net \cite{PUNet}, MPU \cite{MPU}, and PU-GAN \cite{PUGAN}, and also the optimization-based EAR \cite{huang2013edge}. As we have elaborated in Sec. 5.1, the existing deep learning-based methods use training ground truths of output point sets that are \emph{uniformly} sampled from the object surfaces, while our CAD-PU uses curvature-adaptive ones; to enable the comparisons, we have used an evaluation criterion that measures how better the upsampling results obtained by different methods are able to reconstruct the underlying surfaces; we implement the evaluation criterion by first reconstructing the surface mesh from each upsampling result, and then computing the point set distance-based errors between the reconstructed surface and the ground-truth one, where Chamfer distance (CD) and Hausdorff distance (HD) are used as the metrics of point set distance. Under such an evaluation criterion, we have also reported in Sec. 5.2 the results of PU-Net, MPU, and PU-GAN when training them with the same curvature-adaptive output point sets as those used by our CAD-PU.

For an interest of reference, we supplement here the quantitative errors of different methods by \emph{directly} comparing each upsampling result with the ground-truth point set, i.e., the manner of evaluation used in existing methods \cite{PUNet,MPU,PUGAN}. In other words, when the training outputs are uniformly sampled point sets, we compute the CD and HD values between each testing result and the ground truth of uniformly sampled point set; when the training outputs are curvature-adaptively sampled point sets, we compute the CD and HD values between each testing result and the ground truth of curvature-adaptive point set. Table \ref{TableSuppSOTA} reports the quantitative errors of different methods under the two training cases.

\noindent\textbf{Remarks} We emphasize that for the quantitative errors of the two training cases reported in Table \ref{TableSuppSOTA}: (1) they are \emph{not directly comparable} since their ground-truth point sets are different, and (2) they \emph{do not} well reflect the quality of each upsampled point set as an improved approximation of the underlying surface.

\begin{table}[!t]
 \caption{Quantitative errors of different methods by directly computing the Chamfer distance (CD) and Hausdorff distance (HD) between each upsampled point set and its ground truth. We consider two cases whose training ground truths of output point sets are distributed either uniformly or in a curvature-adaptive manner. Results are obtained by averaging over the $27$ testing objects collected in \cite{PUGAN}. }
 \label{TableSuppSOTA}
 \centering
    \begin{tabular}{c|c|cc}
         \hline
         Training ground truth & Methods & CD ($10^{-3}$) & HD ($10^{-3}$) \\
         \hline
         \multirow{3}{*}{Uniform point set}  & PU-Net \cite{PUNet} & $0.72$ & $8.94$\\
                                 & MPU \cite{MPU} & $0.28$ & $\mathbf{2.33}$\\
                                 & PU-GAN \cite{PUGAN} & $\mathbf{0.24}$ & $4.55$ \\
         \hline
         \hline
                           & PU-Net \cite{PUNet} & $0.66$ & $7.42$ \\
         Curvature-adaptive & MPU \cite{MPU} & $0.29$ & $3.13$\\
         point set          & PU-GAN \cite{PUGAN} & $0.27$& $8.04$ \\
                           & Our CAD-PU & $\mathbf{0.26}$ & $\mathbf{2.34}$ \\
         \hline
    \end{tabular}
\end{table}

\end{sloppypar}
\end{document}